\documentclass{article}
\usepackage[utf8]{inputenc}

\usepackage[a4paper, total={6in, 8in}]{geometry}

\usepackage{graphicx}

\usepackage{authblk}
\usepackage[square,numbers]{natbib}
\usepackage{amsmath}
\usepackage{amssymb}
\usepackage{bm}
\usepackage{bbm}
\usepackage{mathtools}
\DeclarePairedDelimiter{\ceil}{\lceil}{\rceil}
\usepackage{todonotes}

\usepackage{multirow}

\usepackage{caption}
\usepackage{subcaption}
\usepackage{hyperref}
\usepackage{booktabs}

\newcommand{\cost}[1]{\operatorname{cost}(#1)}
\newcommand{\stime}[1]{\operatorname{time}(#1)}
\newcommand{\heat}[1]{\operatorname{heat}(#1)}
\newcommand{\potential}[1]{\operatorname{potential}(#1)}
\newcommand{\score}[1]{\operatorname{score}(#1)}
\newcommand{\current}[1]{\operatorname{current}(#1)}
\newcommand{\visited}[1]{\operatorname{visited}(#1)}
\newcommand{\remcap}[1]{\operatorname{capacity}(#1)}
\newcommand{\parent}[1]{\operatorname{parent}(#1)}
\newcommand{\nodepotential}[2]{\operatorname{potential}_{#2}(#1)}
\newcommand{\depot}{\textsc{dep}}
\newcommand{\capacity}{\textsc{capacity}}
\newcommand{\knn}{\textsc{knn}}

\newcommand{\refapp}[1]{Appendix \ref{#1}}

\begin{document}

\title{Deep Policy Dynamic Programming \\ for Vehicle Routing Problems}

\author[1,2]{Wouter Kool\thanks{Corresponding author: 
  \texttt{w.w.m.kool@uva.nl}.}}
\author[1]{Herke van Hoof}
\author[1,2]{Joaquim Gromicho}
\author[1]{Max Welling}
\affil[1]{University of Amsterdam}
\affil[2]{ORTEC}

\date{}

\maketitle              

\begin{abstract}
Routing problems are a class of combinatorial problems with many practical applications. Recently, end-to-end deep learning methods have been proposed to learn approximate solution heuristics for such problems. In contrast, classical dynamic programming (DP) algorithms guarantee optimal solutions, but scale badly with the problem size. We propose \emph{Deep Policy Dynamic Programming} (DPDP), which aims to combine the strengths of learned neural heuristics with those of DP algorithms.
DPDP prioritizes and restricts the DP state space using a policy derived from a deep neural network, which is trained to predict edges from example solutions. We evaluate our framework on the travelling salesman problem (TSP), the vehicle routing problem (VRP) and TSP with time windows (TSPTW) and show that the neural policy improves the performance of (restricted) DP algorithms, making them competitive to strong alternatives such as LKH, while also outperforming most other `neural approaches' for solving TSPs, VRPs and TSPTWs with 100 nodes.
\end{abstract}

\section{Introduction}
Dynamic programming (DP) is a powerful framework for solving optimization problems by solving smaller subproblems through the principle of optimality \citep{bellman1952theory}. Famous examples are Dijkstra's algorithm \citep{dijkstra1959note} for the shortest route between two locations, and the classic Held-Karp algorithm for the travelling salesman problem (TSP) \citep{held1962dynamic,bellman1962dynamic}. Despite their long history, dynamic programming algorithms for vehicle routing problems (VRPs) have seen limited use in practice, primarily due to their bad scaling performance.
More recently, a line of research has attempted the use of machine learning (especially deep learning) to automatically learn heuristics for solving routing problems \citep{vinyals2015pointer,bello2016neural,nazari2018reinforcement,kool2018attention,chen2019learning}. While the results are promising, most learned heuristics are not (yet) competitive to `traditional' algorithms such as LKH \citep{helsgaun2017extension} and lack (asymptotic) guarantees on their performance.

\begin{figure}
     \centering
     \begin{subfigure}[t]{0.3\textwidth}
         \begin{center}
        \centerline{\includegraphics[width=\columnwidth]{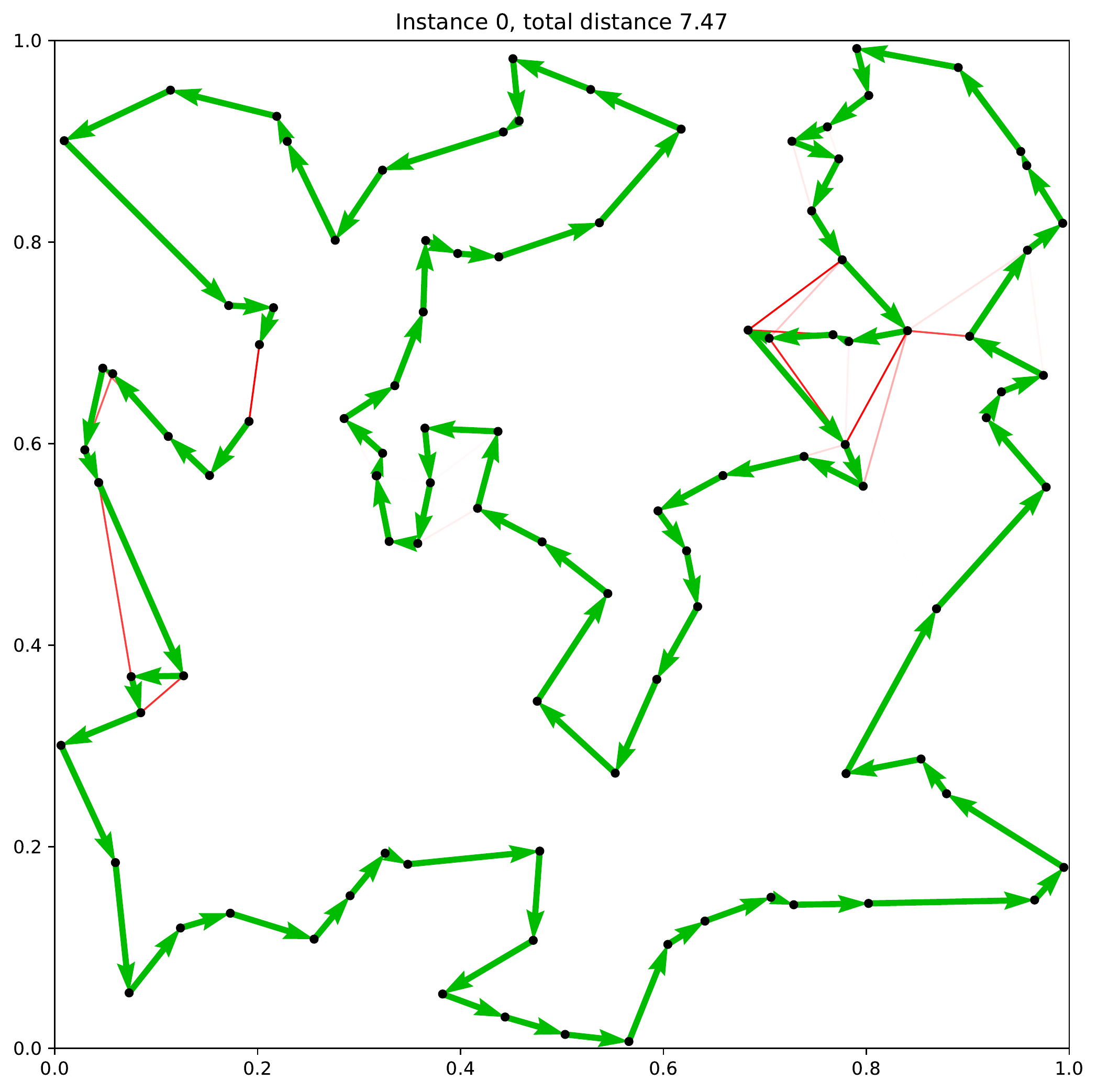}}
        \caption{Travelling Salesman Problem}
        \label{fig:example_tsp}
        \end{center}
     \end{subfigure}
     \hfill
     \begin{subfigure}[t]{0.3\textwidth}
         \begin{center}
        \centerline{\includegraphics[width=\columnwidth]{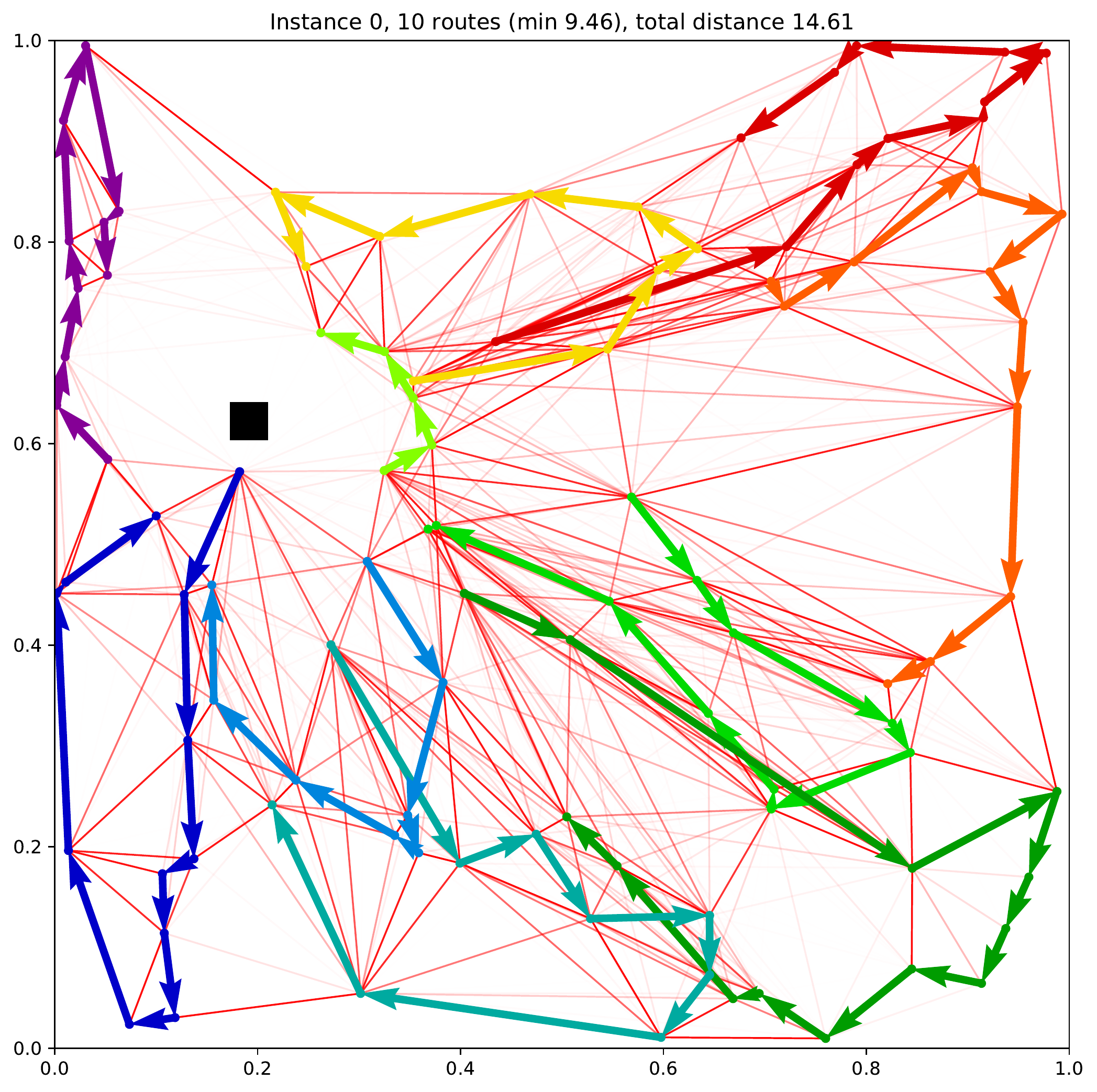}}
        \caption{Vehicle Routing Problem}
        \label{fig:example_vrp}
        \end{center}
     \end{subfigure}
     \hfill
     \begin{subfigure}[t]{0.3\textwidth}
         \begin{center}
        \centerline{\includegraphics[width=\columnwidth]{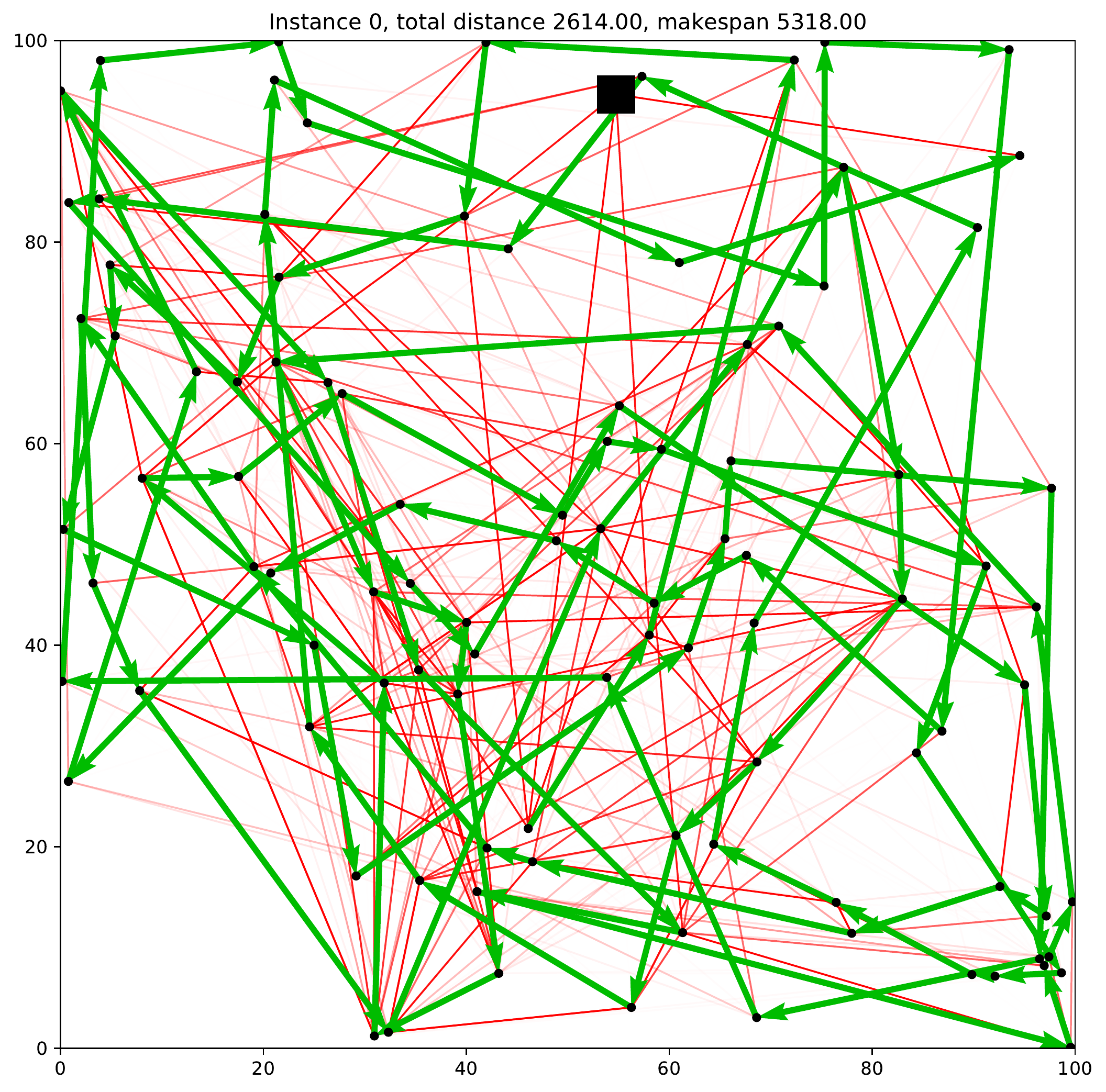}}
        \caption{TSP with Time Windows}
        \label{fig:example_tsptw}
        \end{center}
     \end{subfigure}
        \caption{Heatmap predictions (red) and solutions (colored) by DPDP (VRP depot edges omitted).}
        \label{fig:examples}
\end{figure}

In this paper, we propose \emph{Deep Policy Dynamic Programming} (DPDP) as a framework for solving vehicle routing problems. 
The key of DPDP is to combine the strengths of deep learning and DP, by restricting the DP state space (the search space) using a policy derived from a neural network.
In Figure \ref{fig:examples} it can be seen how the neural network indicates promising parts of the search space (through a \emph{heatmap} over the edges of the graph), which is then used by the DP algorithm to find a good solution. DPDP is more powerful than some related ideas \citep{yang2018boosting,van2019approximate,xu2020deep,cappart2020combining,li2018combinatorial} as it combines supervised training of a large neural network with just a \emph{single} model evaluation at test time, to enable running a large scale guided search using DP.
The DP framework is flexible as it can model a variety of realistic routing problems with difficult practical constraints \citep{gromicho2012restricted}. We illustrate this by testing DPDP on the TSP, the capacitated VRP and the TSP with (hard) time window constraints (TSPTW).

\begin{figure}
\begin{center}
\centerline{\includegraphics[width=\columnwidth,page=5,trim={0.6cm, 4.3cm, 2.1cm, 4.3cm},clip]{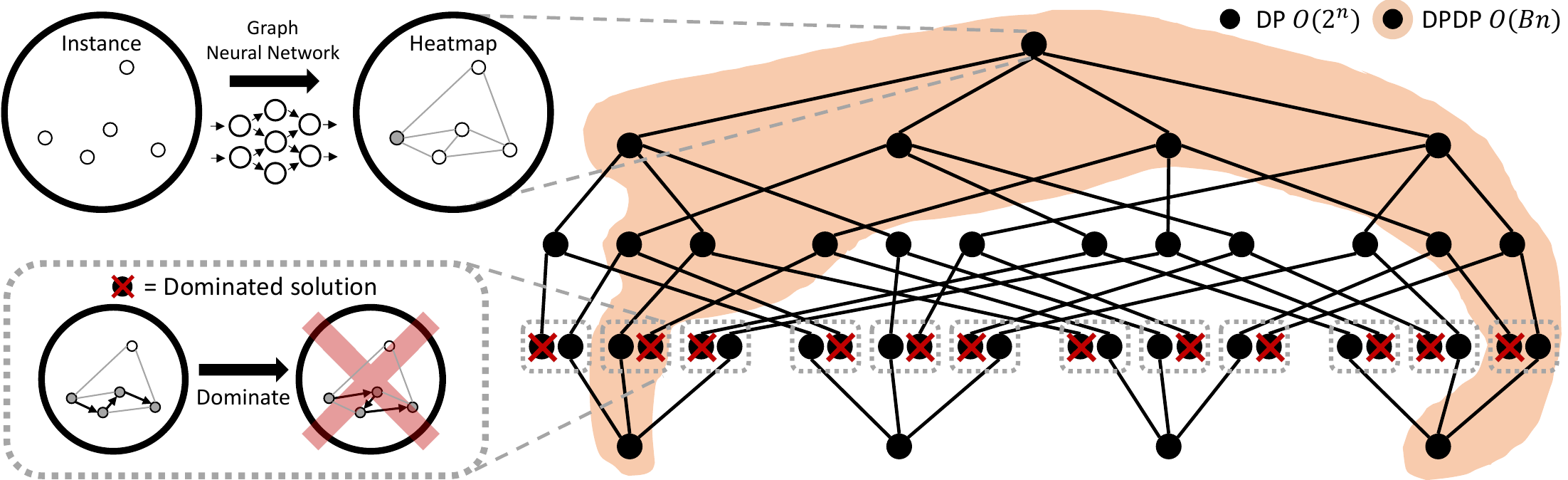}}
\caption{DPDP for the TSP. A GNN creates a (sparse) heatmap indicating promising edges, after which a tour is constructed using forward dynamic programming. In each step, at most $B$ solutions are expanded according to the heatmap policy, restricting the size of the search space. Partial solutions are dominated by shorter (lower cost) solutions with the same DP state: the same nodes visited (marked grey) and current node  (indicated by dashed rectangles). }
\label{fig:DPDP}
\end{center}
\vskip -0.2in
\end{figure}

In more detail, the starting point of our proposed approach is a \emph{restricted dynamic programming} algorithm \citep{gromicho2012restricted}. Such an algorithm heuristically reduces the search space by retaining only the $B$ most promising solutions per iteration. The selection process is very important as it defines the part of the DP state space considered and, thus, the quality of the solution found (see Fig.~\ref{fig:DPDP}).
Instead of manually defining a selection criterion, DPDP defines it using a (sparse) heatmap of promising route segments obtained by pre-processing the problem instance using a (deep) graph neural network (GNN)~\cite{joshi2019efficient}.   
This approach is reminiscent of neural branching policies for branch-and-bound algorithms \citep{gasse2019exact,nair2020solving}.

In this work, we thus aim for a `neural boost' of DP algorithms, by using a GNN for scoring partial solutions. Prior work on `neural' vehicle routing has focused on auto-regressive models \citep{vinyals2015pointer,bello2016neural,deudon2018learning,kool2018attention}, but they have high computational cost when combined with (any form of) search, as the model needs to be evaluated for each partial solution considered. Instead, we use a model to predict a heatmap indicating promising edges \citep{joshi2019efficient}, and define the \emph{score} of a partial solution as the `heat' of the edges it contains (plus an estimate of the `heat-to-go' or \emph{potential} of the solution). As the neural network only needs to be evaluated \emph{once} for each instance, this enables a \emph{much larger search} (defined by $B$), making a good trade-off between quality and computational cost. Additionally, we can apply a threshold to the heatmap to define a sparse graph on which to run the DP algorithm, reducing the runtime by eliminating many solutions.

Figure \ref{fig:DPDP} illustrates DPDP. In Section \ref{sec:experiments}, we show that DPDP significantly improves over `classic' restricted DP algorithms. Additionally, we show that DPDP outperformes most other `neural' approaches for TSP, VRP and TSPTW and is competitive with the highly-optimized LKH solver \citep{helsgaun2017extension} for VRP, while achieving similar results much faster for TSP and TSPTW.
For TSPTW, DPDP also outperforms the best open-source solver we could find \citep{da2010general}, illustrating the power of DPDP to handle difficult hard constraints (time windows).

\section{Related work}
DP has a long history as an exact solution method for routing problems \citep{laporte1992vehicle,toth2014vehicle}, e.g.\ the TSP with time windows \citep{dumas1995optimal} and precedence constraints \citep{mingozzi1997dynamic}, but is limited to small problems due to the curse of dimensionality.
Restricted DP (with heuristic policies) has been used to address, e.g., the time dependent TSP \citep{malandraki1996restricted}, and has been generalized into a flexible framework for VRPs with different types of practical constraints \citep{gromicho2012restricted}. DP approaches have also been shown to be useful in settings with difficult practical issues such as time-dependent travel times and driving regulations \citep{kok2010dynamic} or stochastic demands \citep{novoa2009approximate}. For more examples of DP for routing (and scheduling), see \citep{Hoorn16}. For sparse graphs, alternative, but less flexible, formulations can be used \citep{cook2003tour}.

Despite the flexibility, DP methods have not gained much popularity compared to heuristic search approaches such as R\&R \citep{schrimpf2000record}, ALNS \citep{ropke2006adaptive}, LKH \citep{helsgaun2017extension}, HGS \citep{vidal2012hybrid,vidal2020hybrid} or FILO \citep{accorsi2020fast}, which, while effective, have limited flexibility as special operators are needed for different types of problems. While restricted DP was shown to have superior performance on \emph{realistic} VRPs with many constraints \citep{gromicho2012restricted}, the performance gap of around 10\% for standard (benchmark) VRPs (with time windows) is too large to popularize this approach. We argue that the missing ingredient is a strong but computationally cheap policy for selecting which solutions to considered, which is the motivation behind DPDP.

In the machine learning community, deep neural networks (DNNs) have recently boosted performance on various tasks \citep{lecun2015deep}. After the first DNN model was trained (using example solutions) to construct TSP tours \citep{vinyals2015pointer}, many improvements have been proposed, e.g.\ different training strategies such as reinforcement learning (RL) \citep{bello2016neural,joshi2019learning,delarue2020reinforcement,kwon2020pomo} and model architectures, which enabled the same idea to be used for other routing problems \citep{nazari2018reinforcement,kool2018attention,deudon2018learning,peng2019deep,falkner2020learning,xin2020step,ma2021learning}. Most constructive neural methods are \emph{auto-regressive}, evaluating the model many times to predict one node at the time, but other works have considered predicting a `heatmap' of promising edges \emph{at once} \citep{nowak2017note,joshi2019efficient,fu2020generalize}, which allows a tour to be constructed (using sampling or beam search) without further evaluating the model. An alternative to constructive methods is `learning to search', where a neural network is used to guide a search procedure such as local search \citep{chen2019learning,lu2020learning,gao2020learn,wu2019learning,hottung2019neural,kim2021learning,li2021learning,xin2021neurolkh,hottung2021learning}. Scaling to instances beyond 100 nodes remains challenging \citep{ma2019combinatorial,fu2020generalize}.

The combination of machine learning with DP has been proposed in limited settings \citep{yang2018boosting,van2019approximate,xu2020deep}. Most related to our approach, a DP algorithm for TSPTW, guided by an RL agent, was implemented using an existing solver \citep{cappart2020combining}, which is less efficient than DPDP (see Section \ref{sec:experiments_tsptw}). Also similar to our approach, a neural network predicting edges has been combined with tree search \citep{li2018combinatorial} and local search for maximum independent set (MIS). Whereas DPDP directly builds on the idea of predicting promising edges \citep{li2018combinatorial,joshi2019efficient}, it uses these more efficiently through a policy with \emph{potential function} (see Section \ref{sec:dpdp_tsp}), and by using DP rather than tree search or beam search, we exploit known problem structure in a principled and general manner. As such, DPDP obtains strong performance without using extra heuristics such as local search. For a wider view on machine learning for routing problems and combinatorial optimization, see \citep{mazyavkina2020reinforcement,vesselinova2020learning}.

\section{Deep Policy Dynamic Programming}
DPDP uses an existing graph neural network \citep{joshi2019efficient} (which we modify for VRP and TSPTW) to predict a heatmap of promising edges, which is used to derive the policy for scoring partial solutions in the DP algorithm. The DP algorithm starts with a \emph{beam} of a single initial (empty) solution. It then proceeds by iterating the following steps: (1) all solutions on the beam are expanded, (2) dominated solutions are removed for each \emph{DP state}, (3) the $B$ best solutions according to the scoring policy define the beam for the next iteration. These steps are illustrated in Fig.~\ref{fig:DPDP}. The objective function is used to select the best solution from the final beam.
The resulting algorithm is a \emph{beam search} over the \emph{DP state space} (which is \emph{not} a `standard beam search' over the \emph{solution space}!), where $B$ is the \emph{beam size}. DPDP is asymptotically optimal as using $B = n \cdot 2^n$ for a TSP with $n$ nodes guarantees optimal results, but choosing smaller $B$ allows to trade performance for computational cost.

DPDP is a generic framework that can be applied to different problems, by defining the following ingredients: (1) the \textbf{variables} to track while constructing solutions, (2) the \textbf{initial solution}, (3) \textbf{feasible actions} to expand solutions, (4) rules to define \textbf{dominated solutions} and (5) a \textbf{scoring policy} for selecting the $B$ solutions to keep.
A solution is always (uniquely) defined as a sequence of actions, which allows the DP algorithm to construct the final solution by backtracking. In the next sections, we define these ingredients for the TSP, VRP and TSPTW.

\subsection{Travelling Salesman Problem}
\label{sec:dpdp_tsp}
We implement DPDP for Euclidean TSPs with $n$ nodes on a (sparse) graph, where the cost for edge $(i,j)$ is given by $c_{ij}$, the Euclidean distance between the nodes $i$ and $j$. The objective is to construct a tour that visits all nodes (and returns to the start node) and minimizes the total cost of its edges.

For each partial solution, defined by a sequence of actions $\bm{a}$, the \textbf{variables} we track are $\cost{\bm{a}}$, the total \emph{cost} (distance), $\current{\bm{a}}$, the current node, and $\visited{\bm{a}}$, the set of visited nodes (including the start node). 
Without loss of generality, we let 0 be the start node, so we initialize the beam at step $t = 0$ with the empty \textbf{initial solution} with $\cost{\bm{a}} = 0$, $\current{\bm{a}} = 0$ and $\visited{\bm{a}} = \{0\}$.
At step $t$, the action $a_t \in \{0, ..., n-1\}$ indicates the next node to visit, and is a \textbf{feasible action} for a partial solution $\bm{a} = (a_0, ..., a_{t-1})$ if ($a_{t-1}, a_t)$ is an edge in the graph and $a_t \not \in \visited{\bm{a}}$, or, when all are visited, if $a_t = 0$ to return to the start node. When expanding the solution to $\bm{a}' = (a_0, ..., a_{t})$, we can compute the tracked variables incrementally as:
\begin{equation}
\label{eq:tsp_state_update}
    \cost{\bm{a}'} = \cost{\bm{a}} + c_{\current{a},a_t}, \; \current{\bm{a}'} = a_{t}, \; \visited{\bm{a}'} = \visited{\bm{a}} \cup \{a_t\}.
\end{equation}
A (partial) solution $\bm{a}$ is a \textbf{dominated solution} if there exists a (dominating) solution $\bm{a}^*$ such that
$\visited{\bm{a}^*} = \visited{\bm{a}}$, $\current{\bm{a}^*} = \current{\bm{a}}$ and $\cost{\bm{a}^*} < \cost{\bm{a}}$. We refer to the tuple $(\visited{\bm{a}}, \current{\bm{a}})$ as the \emph{DP state}, so removing all dominated partial solutions, we keep exactly one minimum-cost solution for each unique DP state\footnote{If we have multiple partial solutions with the same state and cost, we can arbitrarily choose one to dominate the other(s), for example the one with the lowest index of the current node.}.
Since a solution can only dominate other solutions with the same set of visited nodes, we only need to remove dominated solutions from sets of solutions with the same number of actions. Therefore, we can efficiently execute the DP algorithm in iterations, where at step $t$ all solutions have (after $t$ actions) $t + 1$ visited nodes (including the start node), keeping the memory need at $O(B)$ states (with $B$ the beam size).

We define the \textbf{scoring policy} using a pretrained model \citep{joshi2019efficient}, which takes as input node coordinates and edge distances to predict a raw `heatmap' value $\hat{h}_{ij} \in (0,1)$ for each edge $(i,j)$. The model was trained to predict optimal solutions, so $\hat{h}_{ij}$ can be seen as the probability that edge $(i,j)$ is in the optimal tour. We force the heatmap to be symmetric thus we define $h_{ij} = \max \{\hat{h}_{ij}, \hat{h}_{ji}\}$. The policy is defined using the heatmap values, in such a way to select the (partial) solutions with the largest total \emph{heat}, while also taking into account the (heat) \emph{potential} for the unvisited nodes. The policy thus selects the $B$ solutions which have the highest \emph{score}, defined as $\score{\bm{a}} = \heat{\bm{a}} + \potential{\bm{a}}$, with $\heat{\bm{a}} = \sum_{i=1}^{t-1} h_{a_{i-1},a_{i}}$, i.e.\ the sum of the heat of the edges, which can be computed incrementally when expanding a solution.
The potential is added as an estimate of the `heat-to-go' (similar to the heuristic in $A^*$ search) for the remaining nodes, and avoids the `greedy pitfall' of selecting the best edges while skipping over nearby nodes, which would prevent good edges from being used later. It is defined as $\potential{\bm{a}} = \nodepotential{\bm{a}}{0} + \sum_{i \not\in \visited{\bm{a}}} \nodepotential{\bm{a}}{i}$ with $\nodepotential{\bm{a}}{i} = w_i \sum_{j \not\in \visited{\bm{a}}} \frac{h_{ji}}{\sum_{k=0}^{n-1} h_{ki}}$,
where $w_i$ is the node \emph{potential weight} given by $w_i = (\max_{j} h_{ji}) \cdot (1 - 0.1( \frac{c_{i0}}{\max_{j} c_{j0}} - 0.5))$.
By normalizing the heatmap values for incoming edges, the (remaining) potential for node $i$ is initially equal to $w_i$ but decreases as good edges become infeasible due to neighbours being visited. The node potential weight $w_i$ is equal to the maximum incoming edge heatmap value (an upper bound to the heat contributed by node $i$), which gets multiplied by a factor 0.95 to 1.05 to give a higher weight to nodes closer to the start node, which we found helps to encourage the algorithm to keep edges that enable to return to the start node. The overall heat + potential function identifies promising partial solutions and is computationally cheap.

\subsection{Vehicle Routing Problem}
For the VRP, we add a special depot node $\depot$ to the graph. Node $i$ has a demand $d_i$, and the goal is to minimize the cost for a set of routes that visit all nodes. Each route must start and end at the depot, and the total demand of its nodes cannot exceed the vehicle capacity denoted by $\capacity$.

Additionally to the TSP \textbf{variables} $\cost{\bm{a}}$, $\current{\bm{a}}$ and $\visited{\bm{a}}$, we keep track of $\remcap{\bm{a}}$, which is the \emph{remaining} capacity in the current route/vehicle. A solution starts at the depot, so we initialize the beam at step $t = 0$ with the empty \textbf{initial solution} with $\cost{\bm{a}} = 0$, $\current{\bm{a}} = \depot$, $\visited{\bm{a}} = \emptyset$ and $\remcap{\bm{a}} = \capacity$. For the VRP, we do not consider visiting the depot as a separate action. Instead, we define $2n$ actions, where $a_t \in \{0, ..., 2n - 1\}$. The actions $0, ..., n - 1$ indicate a \emph{direct} move from the current node to node $a_t$, whereas the actions $n, ..., 2n - 1$ indicate a move to node $a_t - n$ \emph{via the depot}. \textbf{Feasible actions} are those that move to unvisited nodes via edges in the graph and obey the following constraints. For the first action $a_0$ there is no choice and we constrain (for convenience of implementation) $a_0 \in \{n, ..., 2n - 1\}$. A direct move ($a_t < n$) is only feasible if $d_{a_t} \le \remcap{\bm{a}}$ and updates the state similar to TSP but reduces remaining capacity by $d_{a_t}$. A move via the depot is always feasible (respecting the graph edges and assuming $d_i \le \capacity \, \forall i$) as it resets the vehicle $\capacity$ before subtracting demand, but incurs the `via-depot cost' $c_{ij}^{\depot} = c_{i,\depot} + c_{\depot,j}$. When all nodes are visited, we allow a special action to return to the depot. This somewhat unusual way of representing a CVRP solution has desirable properties similar to the TSP formulation: at step $t$ we have exactly $t$ nodes visited, and we can run the DP in iterations, removing dominated solutions at each step $t$.

For VRP, a partial solution $\bm{a}$ is a \textbf{dominated solution} dominated by $\bm{a}^*$ if $\visited{\bm{a}^*} = \visited{\bm{a}}$ and $\current{\bm{a}^*} = \current{\bm{a}}$ (i.e.\ $\bm{a}^*$ corresponds to the same DP state) and $\cost{\bm{a}^*} \le \cost{\bm{a}}$ and $\remcap{\bm{a}^*} \ge \remcap{\bm{a}}$, with \emph{at least one of the two inequalities being strict}.
This means that for each DP state, given by the set of visited nodes and the current node, we do not only keep the (single) solution with lowest cost (as in the TSP algorithm), but keep the complete set of pareto-efficient solutions in terms of cost and remaining vehicle capacity. This is because a higher cost partial solution may still be preferred if it has more remaining vehicle capacity, and vice versa.

For the VRP \textbf{scoring policy}, we modify the model \cite{joshi2019efficient} to include the depot node and demands. The special depot node gets a separate learned initial embedding parameter, and we add additional edge types for connections to the depot, to mark the depot as being special. Additionally, each node gets an extra input (next to its coordinates) corresponding to $d_i / \capacity$ (where we set $d_{\depot} = 0$). Apart from this, the model remains exactly the same\footnote{Except that we do not use the K-nearest neighbour feature \citep{joshi2019efficient} as it contains no additional information.}. The model is trained on example solutions from LKH \citep{helsgaun2017extension} (see Section \ref{sec:experiments_vrp}), which are not optimal, but still provide a useful training signal.
Compared to TSP, the definition of the heat is slightly changed to accommodate for the `via-depot actions' and is best defined incrementally using the `via-depot heat' $h_{ij}^{\depot} = h_{i,\depot} \cdot h_{\depot,j} \cdot 0.1$, where multiplication is used to keep heat values interpretable as probabilities and in the range $(0, 1)$. The additional penalty factor of 0.1 for visiting the depot encourages the algorithm to minimize the number of vehicles/routes.
The initial heat is 0 and when expanding a solution $\bm{a}$ to $\bm{a}'$ using action $a_t$, the heat is incremented with either $h_{\current{\bm{a}},a_t}$ (if $a_t < n$) or $h_{\current{\bm{a}},a_t - n}^{\depot}$ (if $a_t \ge n$). The potential is defined similarly to TSP, replacing the start node 0 by $\depot$.

\subsection{Travelling Salesman Problem with Time Windows}
\label{sec:tsptw}
For the TSPTW, we also have a special depot/start node 0. The goal is to create a single tour that visits each node $i$ in a time window defined by $(l_i, u_i)$, where the travel time from $i$ to $j$ is equal to the cost/distance $c_{ij}$. It is allowed to wait if arrival at node $i$ is before $l_i$, but arrival cannot be after $u_i$. We minimize the total \emph{cost} (\emph{excluding} waiting time), but to minimize \emph{makespan} (including waiting time), we only need to train on different example solutions. Due to the hard constraints, TSPTW is typically considered more challenging than plain TSP, for which every solution is feasible.

The \textbf{variables} we track and \textbf{initial solution} are equal to TSP except that we add $\stime{\bm{a}}$ which is initially 0 ($= l_0$). \textbf{Feasible actions} $a_t \in \{0, ..., n - 1\}$ are those that move to unvisited nodes via edges in the graph such that the arrival time is no later than $u_{a_t}$ and do not directly eliminate the possibility to visit other nodes in time\footnote{E.g., arriving at node $i$ at $t = 10$ is not feasible if node $j$ has $u_j = 12$ and $c_{ij} = 3$.}. Expanding a solution $\bm{a}$ to $\bm{a'}$ using action $a_t$ updates the time as $\stime{\bm{a'}} = \max\{\stime{\bm{a}} + c_{\current{a},a_t}, l_{a_t}\}$.

For each DP state, we keep all efficient solutions in terms of cost and time, so a partial solution $\bm{a}$ is a \textbf{dominated solution} dominated by $\bm{a}^*$ if $\bm{a}^*$ has the same DP state ($\visited{\bm{a}^*} = \visited{\bm{a}}$ and $\current{\bm{a}^*} = \current{\bm{a}}$) and is strictly better in terms of cost and time, i.e.\ $\cost{\bm{a}^*} \le \cost{\bm{a}}$ and $\stime{\bm{a}^*} \le \stime{\bm{a}}$, with \emph{at least one of the two inequalities being strict}.

The model \citep{joshi2019efficient} for the \textbf{scoring policy} is adapted to include the time windows $(l_i, u_i)$ as node features (in the same unit as coordinates and costs), and we use a special embedding for the depot similar to VRP. Due to the time dimension, a TSPTW solution is \emph{directed}, and edge $(i,j)$ may be good whereas $(j,i)$ may be not, so we adapt the model to enable predictions $h_{ij} \neq h_{ji}$ (see \refapp{app:tsptw}). We generated example training solutions using (heuristic) DP with a large beam size, which was faster than LKH. Given the heat predictions, the score (heat + potential) is exactly as for TSP.

\subsection{Graph sparsity}
\label{sec:graph_sparsity}
As described, the DP algorithm can take into account a sparse graph to define feasible expansions. As our problems are defined on sets of nodes rather than graphs, the use of a sparse graph is an artificial design choice, which allows to significantly reduce the runtime but may sacrifice the possibility to find good or optimal tours. We propose two different strategies for defining the sparse graph on which to run the DP: thresholding the heatmap values $h_{ij}$ and using the K-nearest neighbour ($\knn$) graph. By default, we use a (low) heatmap threshold of $10^{-5}$, which rules out most of the edges as the model confidently predicts (close to) 0 for most edges. This is a secondary way to leverage the neural network (independent of the scoring policy), which can be seen as a form of learned \emph{problem reduction} \cite{sun2020generalization}. For symmetric problems (TSP and VRP), we add $\knn$ edges in both directions. For the VRP, we additionally connect each node to the depot (and vice versa) to ensure feasibility.

\subsection{Implementation \& hyperparameters}
We implement DPDP using PyTorch \citep{paszke2017automatic} to leverage GPU computation. For details, see \refapp{app:implementation}. Our code is publicly available.\footnote{\url{https://github.com/wouterkool/dpdp}} DPDP has very few hyperparameters, but the heatmap threshold of $10^{-5}$ and details like the functional form of e.g.\ the scoring policy are `educated guesses' or manually tuned on a few validation instances and can likely be improved. The runtime is influenced by implementation choices which were tuned on a few validation instances.

\section{Experiments}
\label{sec:experiments}

\subsection{Travelling Salesman Problem}

In Table \ref{tab:results_tsp_vrp} we report our main results for DPDP with beam sizes of 10K (10 thousand) and 100K, for the TSP with 100 nodes on a commonly used test set of 10000 instances \citep{kool2018attention}. We report cost and \emph{optimality gap} (see \citep{kool2018attention}) using Concorde \citep{concorde}, LKH \citep{helsgaun2017extension} and Gurobi \citep{gurobi}, as well as recent results of the strongest methods using neural networks (`neural approaches') from literature. Running times for solving 10000 instances \emph{after training} should be taken as rough indications as some are on different machines, typically with 1 GPU or a many-core CPU (8 - 32). The costs indicated with * are not directly comparable due to slight dataset differences \citep{fu2020generalize}. Times for generating heatmaps (if applicable) is reported separately (as the first term) from the running time for MCTS \citep{fu2020generalize} or DP.
DPDP achieves close to optimal results, strictly outperforming the neural baselines achieving better results in less time (except POMO \citep{kwon2020pomo}, see Section \ref{sec:experiments_vrp}).

\begin{table}
\caption{Mean cost, gap and \emph{total time} to solve 10000 TSP/VRP test instances.}
\label{tab:results_tsp_vrp}
\begin{center}
\begin{footnotesize}
\begin{sc}
\begin{tabular}{l|ccc|ccc}
\toprule
Problem & \multicolumn{3}{c|}{TSP100} & \multicolumn{3}{c}{VRP100} \\
Method & Cost & Gap & Time & Cost & Gap & Time \\
\midrule
Concorde \cite{concorde} & 7.765 & 0.000 \% & 6m \\
Hybrid Genetic Search \cite{vidal2012hybrid,vidal2020hybrid} & & & & 15.563 & 0.000 \% & 6h11m \\
Gurobi \cite{gurobi} & 7.776 & 0.151 \% & 31m \\
LKH \cite{helsgaun2017extension} & 7.765 & 0.000 \% & 42m & 15.647 & 0.536 \% & 12h57m \\
\midrule
\scriptsize{GNN Heatmap + Beam Search \cite{joshi2019efficient}} & 7.87 & 1.39 \% & 40m \\
\scriptsize{Learning 2-opt heuristics \cite{da2020learning}} & 7.83 & 0.87 \% & 41m \\
\scriptsize{Merged GNN Heatmap + MCTS \cite{fu2020generalize}} & 7.764* & 0.04 \% & \tiny{4m + 11m} \\
\scriptsize{Attention Model + Sampling \cite{kool2018attention}} & 7.94 & 2.26 \% & 1h & 16.23 & 4.28 \% & 2h \\
\scriptsize{Step-wise Attention Model \cite{xin2020step}} & 8.01 & 3.20 \% & 29s & 16.49 & 5.96 \% & 39s \\
\scriptsize{Attn. Model + Coll. Policies \cite{kim2021learning}} & 7.81 & 0.54 \% & 12h & 15.98 & 2.68 \% & 5h \\
\scriptsize{Learning improv. heuristics \cite{wu2019learning}} & 7.87 & 1.42 \% & 2h & 16.03 & 3.00 \% & 5h \\
\scriptsize{Dual-Aspect Coll. Transformer \cite{ma2021learning}} & 7.77 & 0.09 \% & 5h & 15.71 & 0.94 \% & 9h \\
\scriptsize{Attention Model + POMO \cite{kwon2020pomo}} & 7.77 & 0.14 \% & 1m & 15.76 & 1.26 \% & 2m \\
\scriptsize{NeuRewriter \cite{chen2019learning}} & & & & 16.10 & 3.45 \% & 1h \\
\scriptsize{Dynamic Attn. Model + 2-opt \cite{peng2019deep}} & & & & 16.27 & 4.54 \% & 6h \\
\scriptsize{Neur. Lrg. Neighb. Search \cite{hottung2019neural}} & & & & 15.99 & 2.74 \% & 1h \\
\scriptsize{Learn to improve \cite{lu2020learning}} & & & & 15.57* & - & 4000h \\
\midrule
DPDP 10K & 7.765 & 0.009 \% & \tiny{10m + 16m} & 15.830 & 1.713 \% & \tiny{10m + 50m} \\ 
DPDP 100K & 7.765 & 0.004 \% & \tiny{10m + 2h35m} & 15.694 & 0.843 \% & \tiny{10m + 5h48m} \\
DPDP 1M & & & & 15.627 & 0.409 \% & \tiny{10m + 48h27m} \\
\bottomrule
\end{tabular}
\end{sc}
\end{footnotesize}
\end{center}
\end{table}

\subsection{Vehicle Routing Problem}
\label{sec:experiments_vrp}
For the VRP, we train the model using 1 million instances of 100 nodes, generated according to the distribution described by \cite{nazari2018reinforcement} and solved using one run of LKH \citep{helsgaun2017extension}. We train using a batch size of 48 and a learning rate of $10^{-3}$ (selected as the result of manual trials to best use our GPUs), for (at most) 1500 epochs of 500 training steps (following \cite{joshi2019efficient}) from which we select the saved checkpoint with the lowest validation loss. We use the validation and test sets by \cite{kool2018attention}. 

Table \ref{tab:results_tsp_vrp} shows the results compared to a recent implementation of Hybrid Genetic Search (HGS)\footnote{\url{https://github.com/vidalt/HGS-CVRP}}, a SOTA heuristic VRP solver \citep{vidal2012hybrid,vidal2020hybrid}. HGS is faster and improves around 0.5\% over LKH, which is typically considered the baseline in related work. We present the results for LKH, as well as the strongest neural approaches and DPDP with beam sizes up to 1 million. Some results used 2000 (different) instances \citep{lu2020learning} and cannot be directly compared\footnote{The running time of 4000 hours (167 days) is estimated from 24min/instance \citep{lu2020learning}.}. DPDP outperforms all other neural baselines, except POMO \citep{kwon2020pomo}, which delivers good results very quickly by exploiting symmetries in the problem. However, as it cannot (easily) improve further with additional runtime, we consider this contribution orthogonal to DPDP. DPDP is competitive to LKH (see also Section \ref{sec:beam_sizes}).

\paragraph{More realistic instances}
\label{sec:experiments_uchoa}
We also train the model and run experiments with instances with 100 nodes from a more realistic and challenging data distribution \citep{uchoa2017new}. This distribution, commonly used in the routing community, has greater variability, in terms of node clustering and demand distributions. LKH failed to solve two of the test instances, which is because LKH by default uses a fixed number of routes equal to a lower bound, given by $\ceil[\Big]{ \frac{\sum_{i=0}^{n-1} d_i}{\capacity}}$, which may be infeasible\footnote{For example, three nodes with a demand of two cannot be assigned to two routes with a capacity of three.}. Therefore we solve these instances by rerunning LKH with an unlimited number of allowed routes (which gives worse results, see Section \ref{sec:beam_sizes}).

DPDP was run on a machine with 4 GPUs, but we also report (estimated) runtimes for 1 GPU (1080Ti), and we compare against 16 or 32 CPUs for HGS and LKH. In Table \ref{tab:results_vrp_uchoa} it can be seen that the difference with LKH is, as expected, slightly larger than for the simpler dataset, but still below 1\% for beam sizes of 100K-1M. We also observed a higher validation loss, so it may be possible to improve results using more training data. Nevertheless, finding solutions within 1\% of the specialized SOTA HGS algorithm, and even closer to LKH, is impressive for these challenging instances, and we consider the runtime (for solving 10K instances) acceptable, especially when using multiple GPUs.

\begin{table}
\caption{Mean cost, gap and \emph{total time} to solve 10000 realistic VRP100 instances.}
\label{tab:results_vrp_uchoa}
\vskip 0.15in
\begin{center}
\begin{small}
\begin{sc}
\begin{tabular}{lcccc}
\toprule
Method & Cost & Gap & Time \tiny{(1 GPU or 16 CPUs)} & Time \tiny{(4 GPUs or 32 CPUs)} \\
\midrule
HGS \cite{vidal2012hybrid,vidal2020hybrid} & 18050 & 0.000 \% & 7h53m & 3h56m \\
LKH \cite{helsgaun2017extension} & 18133 & 0.507 \% & 25h32m & 12h46m \\
\midrule
DPDP 10K & 18414 & 2.018 \% & 10m + 50m & 2m + 13m \\
DPDP 100K & 18253 & 1.127 \% & 10m + 5h48m & 2m + 1h27m \\
DPDP 1M & 18168 & 0.659 \% & 10m + 48h27m & 2m + 12h7m \\
\bottomrule
\end{tabular}
\end{sc}
\end{small}
\end{center}
\vskip -0.1in
\end{table}

\subsection{TSP with Time Windows}
\label{sec:experiments_tsptw}
For the TSP with hard time window constraints, we use the data distribution by \citep{cappart2020combining} and use their set of 100 test instances with 100 nodes. These were generated with small time windows, resulting in a small feasible search space, such that even with very small beam sizes, our DP implementation solves these instances optimally, eliminating the need for a policy. Therefore, we also consider a more difficult distribution similar to \citep{da2010general}, which has larger time windows which are more difficult as the feasible search space is larger\footnote{Up to a limit, as making the time windows infinite size reduces the problem to plain TSP.} \citep{dumas1995optimal}. For details, see \refapp{app:tsptw}. For both distributions, we generate training data and train the model exactly as we did for the VRP.

Table \ref{tab:results_tsptw} shows the results for both data distributions, which are reported in terms of the difference to General Variable Neighbourhood Search (GVNS) \citep{da2010general}, the best open-source solver for TSPTW we could find\footnote{\url{https://github.com/sashakh/TSPTW}}, using 30 runs. For the small time window setting, both GVNS and DPDP find optimal solutions for all 100 instances in just 7 seconds (in total, either on 16 CPUs or a single GPU). LKH fails to solve one instance, but finds close to optimal solutions, but around 50 times slower. BaB-DQN* and ILDS-DQN* \citep{cappart2020combining}, methods combining an existing solver with an RL trained neural policy, take around 15 minutes \emph{per instance} (orders of magnitudes slower) to solve most instances to optimality. Due to complex set-up, we were unable to run BaB-DQN* and ILDS-DQN* ourselves for the setting with larger time windows. In this setting, we find DPDP outperforms both LKH (where DPDP is orders of magnitude faster) and GVNS, in both speed and solution quality. This illustrates that DPDP, due to its nature, is especially well suited to handle constrained problems.

\begin{table}[b]
\caption{Mean cost, gap and \emph{total time} to solve TSPTW100 instances.}
\label{tab:results_tsptw}
\begin{center}
\begin{scriptsize}
\begin{sc}
\begin{tabular}{l|cccc|cccc}
\toprule
Problem & \multicolumn{4}{c|}{Small time windows \citep{cappart2020combining} \tiny{(100 inst.)}} & \multicolumn{4}{c}{Large time windows \citep{da2010general} \tiny{(10K inst.)}} \\
Method & Cost & Gap & Fail & Time & Cost & Gap & Fail & Time \\
\midrule
GVNS 30x \cite{da2010general} & 5129.58 & 0.000 \% & & 7s & 2432.112 & 0.000 \% & & 37m15s \\
GVNS 1x \cite{da2010general} & 5129.58 & 0.000 \% & & <1s & 2457.974 & 1.063 \% & & 1m4s \\
LKH 1x \cite{helsgaun2017extension} & 5130.32  & 0.014 \% & 1.00 \% & 5m48s & 2431.404 & -0.029 \% &  & 34h58m \\
\midrule
BaB-DQN* \cite{cappart2020combining} & 5130.51 & 0.018 \% & & 25h \\
ILDS-DQN* \cite{cappart2020combining} & 5130.45 & 0.017 \% & & 25h \\
\midrule
DPDP 10K & 5129.58 & 0.000 \% & & \tiny{6s + 1s} & 2431.143 & -0.040 \% & & \tiny{10m + 8m7s} \\ 
DPDP 100K & 5129.58 & 0.000 \% & & \tiny{6s + 1s} & 2430.880 & - 0.051 \% & & \tiny{10m + 1h16m} \\
\bottomrule
\end{tabular}
\end{sc}
\end{scriptsize}
\end{center}
\end{table}

\subsection{Ablations}

\begin{figure}
     \centering
     \begin{subfigure}[t]{0.32\textwidth}
         \begin{center}
        \centerline{\includegraphics[width=\columnwidth]{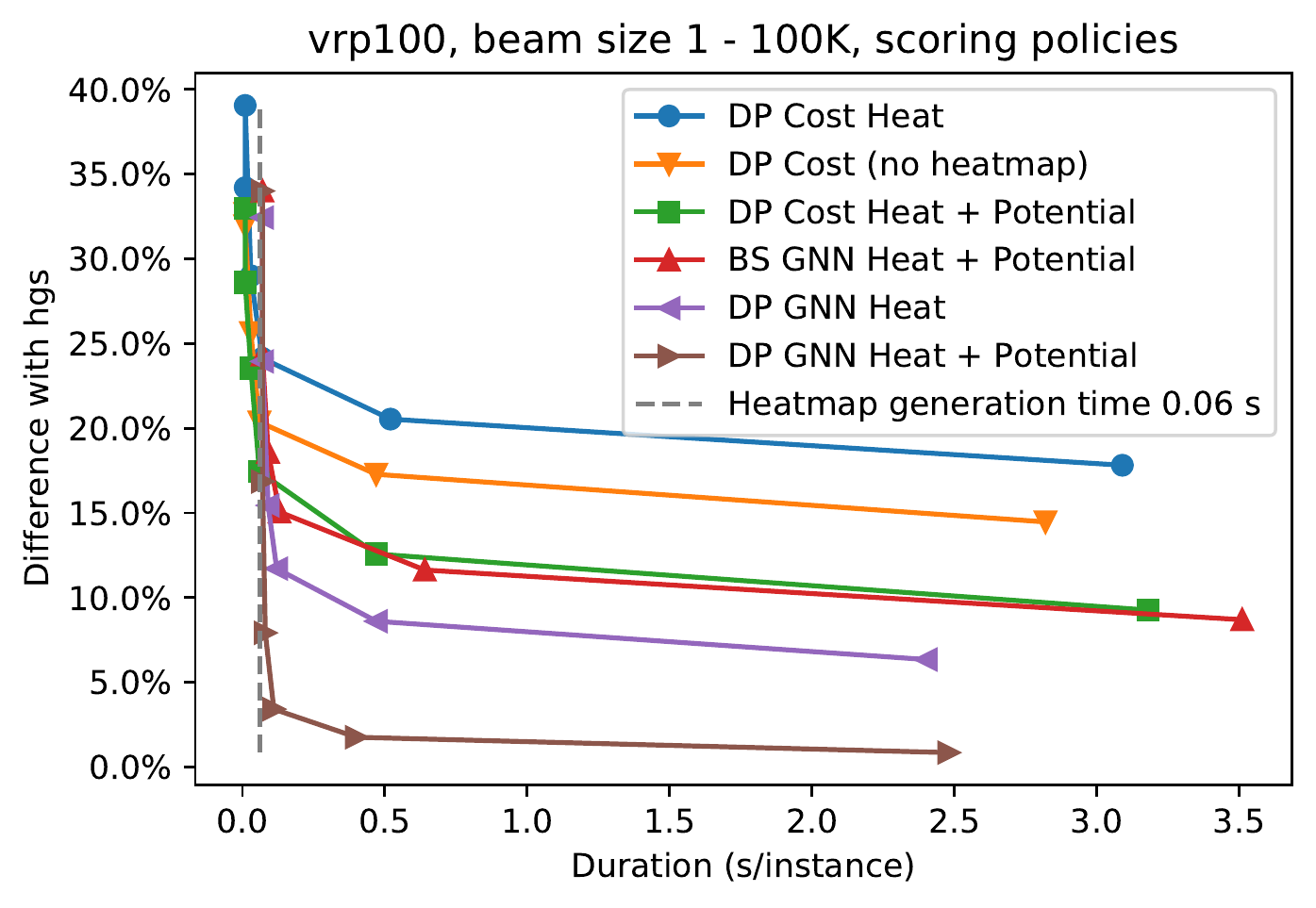}}
        \caption{Different scoring policies, as well as `pure' beam search, for beam sizes 1, 10, 100, 1000, 10K, 100K.}
        \label{fig:scoring_policy}
        \end{center}
     \end{subfigure}
     \hfill
     \begin{subfigure}[t]{0.32\textwidth}
         \begin{center}
        \centerline{\includegraphics[width=\columnwidth]{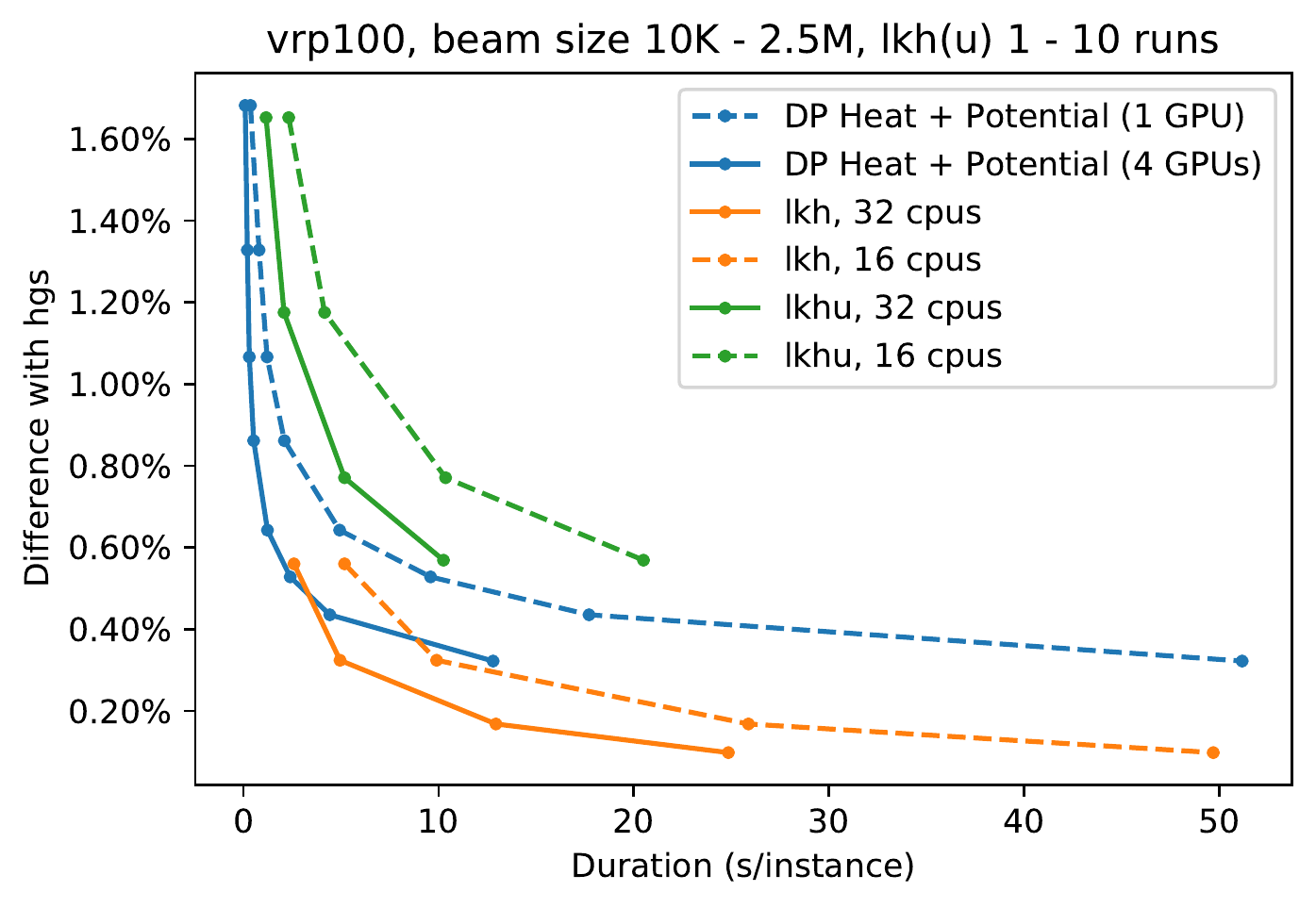}}
        \caption{Beam sizes 10K, 25K, 50K, 100K, 250K, 500K, 1M, 2.5M compared against LKH(U) with 1, 2, 5 and 10 runs.}
        \label{fig:beam_sizes}
        \end{center}
     \end{subfigure}
     \hfill
     \begin{subfigure}[t]{0.32\textwidth}
         \begin{center}
        \centerline{\includegraphics[width=\columnwidth]{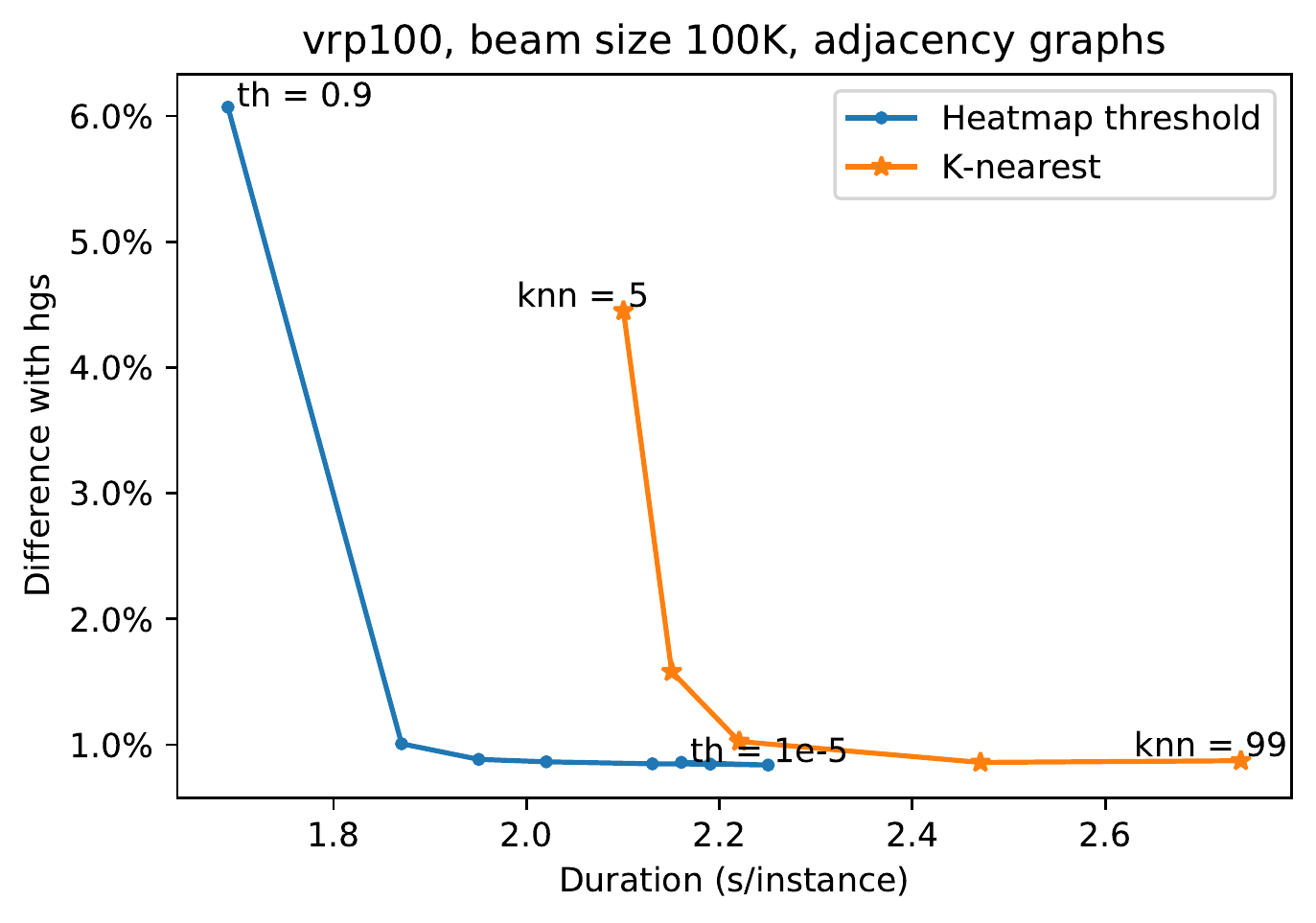}}
        \caption{Sparsities with heatmap thresholds $0.9$, $0.5$, $0.2$, $0.1$, $10^{-2}$, $10^{-3}$, $10^{-4}$, $10^{-5}$ and $\text{knn} = 5$, $10$, $20$, $50$, $99$. Beam size 100K.}
        \label{fig:sparsity}
        \end{center}
     \end{subfigure}
        \caption{DPDP ablations on 100 validation instances of VRP with 100 nodes.}
        \label{fig:three graphs}
\end{figure}

\paragraph{Scoring policy}
\label{sec:scoring_policy}
To evaluate the value of different components of DPDP's \textbf{GNN Heat + Potential} scoring policy, we compare against other variants. \textbf{GNN Heat} is the version without the potential, whereas \textbf{Cost Heat + Potential} and \textbf{Cost Heat} are variants that use a `heuristic' $\hat{h}_{ij} = \frac{c_{ij}}{\max_{k} c_{ik}}$ instead of the GNN. \textbf{Cost} directly uses the current cost of the solution, and can be seen as `classic' restricted DP. Finally, \textbf{BS GNN Heat + Potential} uses beam search without dynamic programming, i.e.\ without removing dominated solutions.
To evaluate only the scoring policy, each variant uses the fully connected graph ($\text{knn} = n - 1$).
Figure \ref{fig:scoring_policy} shows the value of DPDP's potential function, although even without it results are still significantly better than `classic' heuristic DP variants using cost-based scoring policies.
Also, it is clear that using DP significantly improves over a standard beam search (by removing dominated solutions). 
Lastly, the figure illustrates how the time for generating the heatmap using the neural network, despite its significant value, only makes up a small portion of the total runtime.

\paragraph{Beam size}
\label{sec:beam_sizes}
DPDP allows to trade off the performance vs.\ the runtime using the beam size $B$ (and to some extent the graph sparsity, see Section \ref{sec:experiment_sparsity}).
We illustrate this trade-off in Figure \ref{fig:beam_sizes}, where we evaluate DPDP on 100 validation instances for VRP, with different beam sizes from 10K to 2.5M. We also report the trade-off curve for the LKH(U), which is the strongest baseline that can also solve different problems. We vary the runtime using 1, 2, 5 and 10 runs (returning the best solution). LKHU(nlimited) is the version which allows an unlimited number of routes (see Section \ref{sec:experiments_uchoa}).
It is hard to compare GPU vs CPU, so we report (estimated) runtimes for different hardware, i.e.\ 1 or 4 GPUs (with 3 CPUs per GPU) and 16 or 32 CPUs.
We report the difference (i.e.\ the gap) with HGS, analog to how results are reported in Table \ref{tab:results_tsp_vrp}. We emphasize that in most related work (e.g.\ \cite{kool2018attention}), the strongest baseline considered is one run of LKH, so we compare against a much stronger baseline. Also, our goal is not to outperform HGS (which is SOTA and specific to VRP) or LKH, but to show DPDP has reasonable performance, while being a flexible framework for other (routing) problems.

\paragraph{Graph sparsity}
\label{sec:experiment_sparsity}
We test the two graph sparsification strategies described in Section \ref{sec:graph_sparsity} as another way to trade off performance and runtime of DPDP.
In Figure \ref{fig:sparsity}, we experiment with different heatmap thresholds from $10^{-5}$ to 0.9 and different values for $\knn$ from 5 to 99 (fully connected). The heatmap threshold strategy clearly outperforms the $\knn$ strategy as it yields the same results using sparser graphs (and lower runtimes). This illustrates that the heatmap threshold strategy is more informed than the $\knn$ strategy, confirming the value of the neural network predictions.

\section{Discussion}
In this paper we introduced Deep Policy Dynamic Programming, which combines machine learning and dynamic programming for solving vehicle routing problems. The method yields close to optimal results for TSPs with 100 nodes and is competitive to the highly optimized LKH \citep{helsgaun2017extension} solver for VRPs with 100 nodes. On the TSP with time windows, DPDP also outperforms LKH, being significanlty faster, as well as GVNS \citep{da2010general}, the best open source solver we could find. Given that DPDP was not specifically designed for TSPTW, and thus can likely be improved, we consider this an impressive and promising achievement.

The constructive nature of DPDP (combined with search) allows to efficiently address hard constraints such as time windows, which are typically considered challenging in neural combinatorial optimization \citep{bello2016neural,kool2018attention} and are also difficult for local search heuristics (as they need to maintain feasibility while adapting a solution). Given our results on TSP, VRP and TSPTW, and the flexibility of DP as a framework, we think DPDP has great potential for solving many more variants of routing problems, and possibly even other problems that can be formulated using DP (e.g.\ job shop scheduling \citep{gromicho2012solving}).
We hope that our work brings machine learning research for combinatorial optimization closer to the operations research (especially vehicle routing) community, by combining machine learning with DP and evaluating the resulting new framework on different data distributions used by different communities \citep{nazari2018reinforcement,uchoa2017new,cappart2020combining,da2010general}. 

\paragraph{Scope, limitations \& future work}
Deep learning for combinatorial optimization is a recent research direction, which could significantly impact the way practical optimization problems get solved in the future. Currently, however, it is still hard to beat most SOTA problem specific solvers from the OR community. Despite our success for TSPTW, DPDP is not yet a practical alternative in general, but we do consider our results as highly encouraging for further research.
We believe such research could yield significant further improvement by addressing key current limitations: (1) the scalability to larger instances, (2) the dependency on example solutions and (3) the heuristic nature of the scoring function. First, while 100 nodes is not far from the size of common benchmarks (100 - 1000 for VRP \citep{uchoa2017new} and 20 - 200 for TSPTW \citep{da2010general}), scaling is a challenge, mainly due to the `fully-connected' $O(n^2)$ graph neural network. Future work could reduce this complexity following e.g. \citep{lee2019set}. The dependency on example solutions from an existing solver also becomes more prominent for larger instances, but could potentially be removed by `bootstrapping' using DP itself as we, in some sense, have done for TSPTW (see Section \ref{sec:tsptw}). Future work could iterate this process to train the model `tabula rasa' (without example solutions), where DP could be seen analogous to MCTS in \emph{AlphaZero} \citep{silver2018general}. Lastly, the heat + potential score function is a well-motivated but heuristic function that was manually designed as a function of the predicted heatmap. While it worked well for the three problems we considered, it may need suitable adaption for other problems. Training this function end-to-end \citep{daume2005learning,wiseman2016sequence}, while keeping a low computational footprint, would be an interesting topic for future work. 


\clearpage

\bibliographystyle{plain}
\bibliography{references}

\clearpage
\appendix
\section{Implementation}
\label{app:implementation}
We implement the dynamic programming algorithm on the GPU using PyTorch \citep{paszke2017automatic}. While mostly used as a Deep Learning framework, it can be used to speed up generic (vectorized) computations.

\subsection{Beam variables}
For each solution in the beam, we keep track of the following variables (storing them for all solutions in the beam as a vector): the cost, current node, visited nodes and (for VRP) the remaining capacity or (for TSPTW) the current time. As explained, these variables can be computed incrementally when generating expansions. Additionally, we keep a variable vector \emph{parent}, which, for each solution in the current beam, tracks the index of the solution in the previous beam that generated the expanded solution. To compute the score of the policy for expansions efficiently, we also keep track of the score for each solution and the potential for each node for each solution incrementally.

We do not keep past beams in memory, but at the end of each iteration, we store the vectors containing the parents as well as last actions for each solution on the \emph{trace}. As the solution is completely defined by the sequence of actions, this allows to backtrack the solution after the algorithm has finished. To save GPU memory (especially for larger beam sizes), we store the $O(Bn)$ sized trace on the CPU memory.

\begin{figure}
     \centering
     \begin{subfigure}[t]{0.59\textwidth}
    \begin{center}
    \centerline{\includegraphics[width=\columnwidth,page=4,trim={12.4cm, 4.8cm, 12.0cm, 5.5cm},clip]{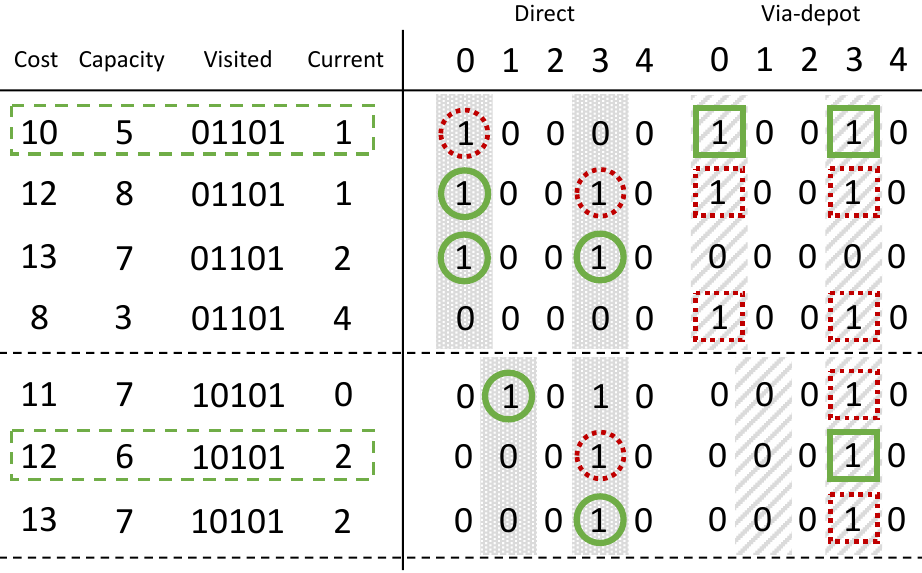}}
    \caption{Example beam for VRP with variables, grouped by set of visited nodes (left) and feasible, non-dominated expansions (right), with $2n$ columns corresponding to $n$ direct expansions and $n$ via-depot expansions. Some expansions to unvisited nodes are infeasible, e.g.\ due to the capacity constraint or a sparse adjacency graph. The shaded areas indicate groups of candidate expansions among which dominances should be checked: for each set of visited nodes there is only one non-dominated via-depot expansion (indicated by solid green square), which must necessarily be an expansion of the solution that has the lowest cost to return to the depot (indicated by the dashed green rectangle ; note that the cost displayed excludes the cost to return to the depot). Direct expansions can be dominated (indicated by red dotted circles) by the single non-dominated via-depot expansion or other direct expansions with the same DP state (set of visited nodes and expanded node, as indicated by the shaded areas). See also Figure \ref{fig:dominate_vrp} for (non-)dominated expansions corresponding to the same DP state.}
    \label{fig:beam_vrp}
    \end{center}
    \end{subfigure}
     \hfill
     \begin{subfigure}[t]{0.375\textwidth}
    \begin{center}
    \centerline{\includegraphics[width=\columnwidth,page=3,trim={12.4cm, 2.8cm, 13.5cm, 2.5cm},clip]{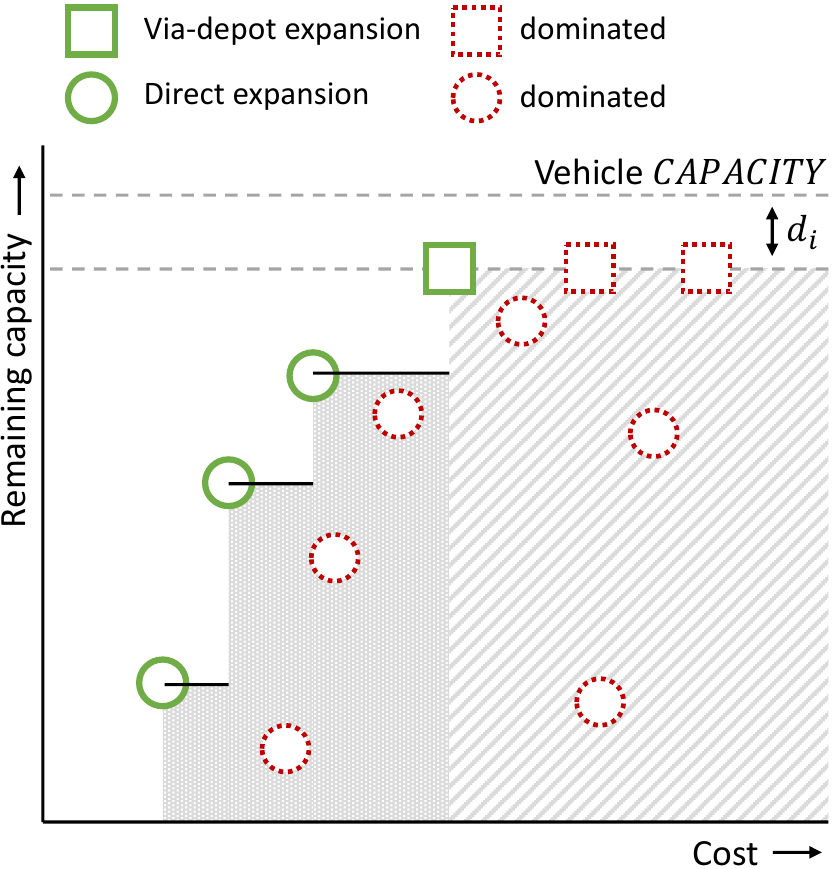}}
    \caption{Example of a set of dominated and non-dominated expansions (direct and via-depot) corresponding to the same DP state (set of visited nodes and expanded node $i$) for VRP. Non-dominated expansions have lower cost or higher remaining capacity compared to all other expansions. The right striped area indicates expansions dominated by the (single) non-dominated via-depot expansion. The left (darker) areas are dominated by individual direct expansions. Dominated expansions in this area have remaining capacity lower than the cumulative maximum remaining capacity when going from left to right (i.e.\ in sorted order of increasing cost), indicated by the black horizontal lines.}
    \label{fig:dominate_vrp}
    \end{center}
    \vskip -0.2in
    \end{subfigure}
    \caption{Implementation of DPDP for VRP}
\end{figure}

For efficiency, we keep the set of visited nodes as a bitmask, packed into 64-bit long integers (2 for 100 nodes). Using bitwise operations with the packed adjacency matrix, this allows to quickly check feasible expansions (but we need to \emph{unpack} the mask into boolean vectors to find all feasible expansions explicitly). Figure \ref{fig:beam_vrp} shows an example of the beam (with variables related to the policy and backtracking omitted) for the VRP.

\subsection{Generating non-dominated expansions}
A solution $\bm{a}$ can only dominate a solution $\bm{a}'$ if $\visited{\bm{a}} = \visited{\bm{a}'}$ and $\current{\bm{a}} = \current{\bm{a}'}$, i.e.\ if they correspond to the same \emph{DP state}. If this is the case, then, if we denote by $\parent{\bm{a}}$ the parent solution from which $\bm{a}$ was expanded, it holds that 
\begin{align*}
    \visited{\parent{\bm{a}}} &= \visited{\bm{a}} \setminus \{\current{\bm{a}}\} \\
    &= \visited{\bm{a}'} \setminus \{\current{\bm{a}'}\} \\
    &= \visited{\parent{\bm{a}'}}.
\end{align*}
This means that only expansions from solutions with the same set of visited nodes can dominate each other, so we only need to check for dominated solutions among groups of expansions originating from parent solutions with the same set of visited nodes. Therefore, before generating the expansions, we group the current beam (the parents of the expansions) by the set of visited nodes (see Figure \ref{fig:beam_vrp}). This can be done efficiently, e.g.\ using a lexicographic sort of the packed bitmask representing the sets of visited nodes\footnote{For efficiency, we use a custom function similar to \textsc{torch.unique}, and argsort the returned inverse after which the resulting permutation is applied to all variables in the beam.}.

\subsubsection{Travelling Salesman Problem}
For TSP, we can generate (using boolean operations) the $B \times n$ matrix with boolean entries indicating feasible expansions (with $n$ action columns corresponding to $n$ nodes, similar to the $B \times 2n$ matrix for VRP in Figure \ref{fig:beam_vrp}), i.e.\ nodes that are unvisited and adjacent to the current node. If we find positive entries sequentially for each column (e.g.\ by calling $\textsc{torch.nonzero}$ on the transposed matrix), we get all expansions grouped by the combination of action (new current node) and parent set of visited nodes, i.e.\ grouped by the DP state. We can then trivially find the segments of consecutive expansions corresponding to the same DP state, and we can efficiently find the minimum cost solution for each segment, e.g.\ using \textsc{torch\_scatter} \footnote{\url{https://github.com/rusty1s/pytorch\_scatter}}.

\subsubsection{Vehicle Routing Problem}
For VRP, the dominance check has two dimensions (cost \emph{and} remaining capacity) and additionally we need to consider $2n$ actions: $n$ direct and $n$ via the depot (see Figure \ref{fig:beam_vrp}). Therefore, as we will explain, we check dominances in two stages: first we find (for each DP state) the \emph{single} non-dominated `via-depot' expansion, after which we find all non-dominated `direct' expansions (see Figure \ref{fig:dominate_vrp}).

The DP state of each expansion is defined by the expanded node (the new current node) and the set of visited nodes. For each DP state, there can be only \emph{one}\footnote{Unless we have multiple expansions with the same costs, in which case can pick one arbitrarily.} non-dominated expansion where the last action was via the depot, since all expansions resulting from `via-depot actions' have the same remaining capacity as visiting the depot resets the capacity (see Figure \ref{fig:dominate_vrp}). To find this expansion, we first find, for each unique set of visited nodes in the current beam, the solution that can return to the depot with lowest total cost (thus including the cost to return to the depot, indicated by a dashed green rectangle in Figure \ref{fig:beam_vrp}). The single non-dominated `via-depot expansion' for each DP state must necessarily be an expansion of this solution. Also observe that this via-depot solution cannot be dominated by a solution expanded using a direct action, which will always have a lower remaining vehicle capacity (assuming positive demands) as can bee seen in Figure \ref{fig:dominate_vrp}. We can thus generate the non-dominated via-depot expansion for each DP state efficiently and independently from the direct expansions.

For each DP state, all \emph{direct} expansions with cost higher (or equal) than the via-depot expansion can directly be removed since they are dominated by the via-depot expansion (having higher cost and lower remaining capacity, see Figure \ref{fig:dominate_vrp}). After that, we sort the remaining (if any) direct expansions for each DP state based on the cost (using a segmented sort as the expansions are already grouped if we generate them similarly to TSP, i.e.\ per column in Figure \ref{fig:beam_vrp}). For each DP state, the lowest cost solution is never dominated. The other solutions should be kept only if their remaining capacity is strictly larger than the largest remaining capacity of all lower-cost solutions corresponding to the same DP state, which can be computed using a (segmented) cumulative maximum computation (see Figure 
\ref{fig:dominate_vrp}).

\subsubsection{TSP with Time Windows}
For the TSPTW, the dominance check has two dimensions: cost and time. Therefore, it is similar to the check for non-dominated direct expansions for the VRP (see Figure \ref{fig:dominate_vrp}), but replacing remaining capacity (which should be maximized) by current time (to be minimized). In fact, we could reuse the implementation, if we replace remaining capacity by time multiplied by $-1$ (as this should be minimized). This means that we sort all expansions for each DP state based on the cost, keep the first solution and keep other solutions only if the time is strictly lower than the lowest current time for all lower-cost solutions, which can be computed using a cumulative minimum computation.

\subsection{Finding the top $B$ solutions}
We may generate all `candidate' non-dominated expansions and then select the top $B$ using the score function. Alternatively, we can generate expansions in batches, and keep a streaming top $B$ using a priority queue. We use the latter implementation, where we can also derive a bound for the score as soon as we have $B$ candidate expansions. Using this bound, we can already remove solutions before checking dominances, to achieve some speedup in the algorithm.\footnote{This may give slightly different results if the scoring function is inconsistent with the domination rules, i.e.\ if a better scoring solution would be dominated by a worse scoring solution but is not since that solution is removed using the score bound before checking the dominances.}

\subsection{Performance improvements}
There are many possibilities for improving the speed of the algorithm. For example, PyTorch lacks a segmented sort so we use a much slower lexicographic sort instead. Also an efficient GPU priority queue would allow much speedup, as we currently use sorting as PyTorch' top-$k$ function is rather slow for large $k$. In some cases, a binary search for the $k$-th largest value can be faster, but this introduces undesired CUDA synchronisation points.
We currently use multiprocessing to solve multiple instances on a single GPU in parallel, introducing a lot of Python overhead. A batched implementation would give a significant speedup.

\clearpage
\section{TSP with Time Windows}
\label{app:tsptw}
This section contains additional information for the TSPTW.

\subsection{Adaption of model for TSPTW}
The model updates the edge embedding $e_{ij}^{l}$ for edge $(i, j)$ in layer $l + 1$ using node embeddings $x_i^l$ and $x_j^l$ with the following equation (Equation (5) in \cite{joshi2019efficient}):
\begin{equation}
    e_{ij}^{l+1} + \text{ReLU}(\text{BatchNorm}(W_3^l e_{ij}^l + W_4^l x_i^l + W_5^l x_j^l))
\end{equation}
where $W_3^l, W_4^l$ and $W_5^l$ are trainable parameters. We found out that the implementation\footnote{\url{https://github.com/chaitjo/graph-convnet-tsp/blob/master/models/gcn\_layers.py}} actually shares the parameters $W_4^l$ and $W_5^l$, i.e.\ $W_4^l = W_5^l$, resulting in $e_{ij}^l = e_{ji}^l$ for all layers $l$, as for $l = 0$ both directions are initialized the same. To allow the model to make different predictions for different directions, we implement $W_5^l$ as a separate parameter, such that the model can have different representations for edges $(i, j)$ and $(j, i)$. We define the training labels accordingly for directed edges, so if edge $(i, j)$ is in the directed solution, it will have a label 1 whereas the edge $(j, i)$ will not (for the undirected TSP and VRP, both labels are 1).

\subsection{Dataset generation}
We found that using our DP formulation for TSPTW, the instances by \cite{cappart2020combining} were all solved optimally, even with a very small beam size (around 10). This is because there is very little overlap in the time windows as a result of the way they are generated, and therefore very few actions are feasible as most of the actions would `skip over other time windows' (advance the time so much that other nodes can no longer be served)\footnote{If all time windows are disjoint, there is only one feasible solution. Therefore, the amount of overlap in time windows determines to some extent the `branching factor' of the problem and the difficulty.}. We conducted some quick experiments with a weaker DP formulation, that only checks if actions \emph{directly} violate time windows, but does not check if an action causes other nodes to be no longer reachable in their time windows. Using this formulation, the DP algorithm can run into many dead ends if just a single node gets skipped, and using the GNN policy (compared to a cost based policy as in Section \ref{sec:scoring_policy}) made the difference between good solutions and no solution at all being found.

We made two changes to the data generation procedure by \cite{cappart2020combining} to increase the difficulty and make it similar to \cite{da2010general}, defining the `large time window' dataset. First, we sample the time windows around arrival times when visiting nodes in a random order without any waiting time, which is different from \cite{cappart2020combining} who `propagate' the waiting time (as a result of time windows sampled). Our modification causes a tighter schedule with more overlap in time windows, and is similar to \cite{da2010general}. Secondly, we increase the maximum time window size from 100 to 1000, which makes that the time windows are in the order of 10\% of the horizon\footnote{Serving 100 customers in a 100x100 grid, empirically we find the total schedule duration including waiting (the makespan) is around 5000.}. This doubles the maximum time window size of 500 used by \cite{da2010general} for instances with 200 nodes, to compensate for half the number of nodes that can possibly overlap the time window.

To generate the training data, for practical reasons we used DP with the heuristic `cost heat + potential' strategy and a large beam size (1M), which in many cases results in optimal solutions being found.

\end{document}